# MN-Pair Contrastive Damage Representation and Clustering for Prognostic Explanation

Takato Yasuno[1], Masahiro Okano[1] and Junichiro Fujii[1]

[1]Yachiyo Engineering, Co.,Ltd. Research Institute for Infrastructure Paradigm Shift, Japan.
{tk-yasuno, ms-okano, jn-fujii}@yachiyo-eng.co.jp

**Abstract –** For infrastructure inspections, damage representation does not constantly match the predefined classes of damage grade, resulting in detailed clusters of unseen damages or more complex clusters from overlapped space between two grades. The damage representation has fundamentally complex features; consequently, not all the damage classes can be perfectly predefined. The proposed MN-pair contrastive learning method helps to explore an embedding damage representation beyond the predefined classes by including more detailed clusters. It maximizes both the similarity of M-1 positive images close to an anchor and dissimilarity of N-1 negative images using both weighting loss functions. It learns faster than the N-pair algorithm using one positive image. We proposed a pipeline to obtain the damage representation and used a density-based clustering on a 2-D reduction space to automate finer cluster discrimination. We also visualized the explanation of the damage feature using Grad-CAM for MN-pair damage metric learning. We demonstrated our method in three experimental studies: steel product defect, concrete crack, and the effectiveness of our method and discuss future works.

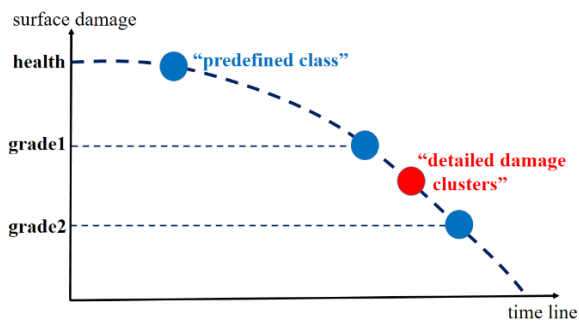

Figure 1. Predefined classes and finer clusters for prognostic damage understanding

## 1 Introduction

### 1.1 Objective and Pipeline

#### 1.1.1 Damage Inspection and Clustering Problem

It is crucial for infrastructure managers to maintain high standards, to ensure user satisfaction during daily operations. Surveillance cameras and drone inspections have enabled progress toward automating the inspection of damaged features and assessing the deterioration of infrastructure. By providing a pair of raw images and damage class labels, we can train the supervised learning to identify a predefined grade of damage and displacement. However, such a damage representation does not constantly match the predefined classes of damage grade, hence, there may be some detailed clusters from the unseen damage spaces or more complex clusters from overlapped spaces between two damage grades, as shown in Figure 1. Damage representations have fundamentally complex features; consequently, not all the damage classes can be perfectly predefined.

In practice, some grades of damage that includes complex features may be occasionally scattered for human annotators to consistently categorize. Furthermore, there are rare occurrences of minor classes of deterioration, making it difficult to establish predefined classes that encompass all possible damage types based solely on our prior knowledge. Contrastive learning methods present an opportunity to explore the embedding damage representations beyond these predefined classes, allowing for the identification of more detailed clusters.

#### 1.1.2 Contrastive Damage Metric Learning

Deep metric learning has been extensively investigated using shared-parameter networks, such as Siamese and triplet networks [1]. Since 2013, various contrastive loss function minimizations have been proposed for learning representations from pairs of similar/dissimilar samples [2]. For face recognition, verification, and clustering, several deep convolutional networks have been used to map from facial images onto a compact Euclidean space, where the distance points indicates the degree of facial similarity [3][4]. The usefulness of deep metric learning for recognizing the target of damaged surface similarity in infrastructure, in the construction domain, has still not been sufficiently understood.

During the lifecycle of a deterioration process, the damaged surface of infrastructures frequently exhibits complex representations. While inspecting against the surface on the parts of a structure using vision-based technology, predefined qualitative grading classes are used to recognize and record any damage observed. Despite the possibility of encountering unseen circumstances and ambiguous statuses that fall outside of predefined classes, further experience may be necessary. Furthermore, complex defects can include multiple types of damages and a mixture of defects that are not typically observed in real-world structural monitoring.

Measuring the similarity of damage in high-dimensional visual images requires the learning of an embedding in a lower-dimensional damaged space. Once such a damage-embedding space has been produced, vision-based inspection tasks such as damage recognition, diagnosis, and prognostic clustering can be easily implemented as standard techniques with damage embedding as feature vectors.

### 1.1.3 Damage Representation Clustering Pipeline

As shown in Figure 2, we proposed a damage representation pipeline for learning damage features and for clustering more detailed heterogeneous damages. The first training stage involved developing a damage metric learning. Our proposed MN-pair contrastive learning method maximizes both the similarity of M-1 positive images to an anchor and the dissimilarity of N-1 negative images, using a weighting MN-pair loss function. Triplet networks share the parameters under a convolutional neural network (CNN). The outputs of the damage-embedding feature set with 16 elements were reduced to two dimensions using the t-SNE algorithm. The second stage was density-based clustering toward more detailed damage clusters with an outlier option using the density-based spatial clustering application with noise (DBSCAN) algorithm. The third explanation stage involved sorting the nearest feature scores and visualizing the tiles of the ten nearest images. Furthermore, each image enabled us to compute heat maps using an adapted Grad-CAM algorithm for the metric learning of MN-pair damage.

## 2 Damage Representation Learning

### 2.1 MN-pair Damage Contrastive Learning

Let $x$ denote the set of inspected input images with predefined deterioration classes. Let $e_i = f(x_i; \theta) \in R^L$ be the $i$–th damage embedding of input $i \in \{1, ..., n\}$ that preserves the damage semantic aspects. Here, $n$ is the number of input images. Furthermore, $\theta$ is a shared parameters under a CNN for damage metric learning, and $L$ is the dimension of the damage embedding space. Let $F_i = e_i/\|e_i\|_2$ be the $l_2$–normalized version. The damage similarity can be measured from the distance between the two images $i_1$ and $i_2$ using the normalized cosine similarity:

$$s_\theta(x_{i_1}, x_{i_2}) = (F_{i_1})^T F_{i_2} \quad (1)$$

where larger values indicate greater similarities. Here, the suffix $T$ denotes the transposed operation.

The N-pair loss approach [5] creates a multiclass classification in which we create a set of N-1 negative $\{x_k^-\}_{k=1}^{N-1}$ and one positive $x_j^+$ for every anchor image $x_i$. We define the following N-pair loss function for each set:

$$L_{N-pair}(\theta; x_i, x_j^+, \{x_k^-\}_{k=1}^{N-1}) = \log\left(1 - \exp\left(s_\theta(x_i, x_j^+)\right) + \sum_{k=1}^{N-1} \exp(s_\theta(x_i, x_k^-))\right) \quad (2)$$

For simple expression, we denote cosine similarities as:

$$s_{i,j}^{a,+} = s_\theta(x_i, x_j^+)/\tau, \; s_{i,k}^{a,-} = s_\theta(x_i, x_k^-)/\tau \quad (3)$$

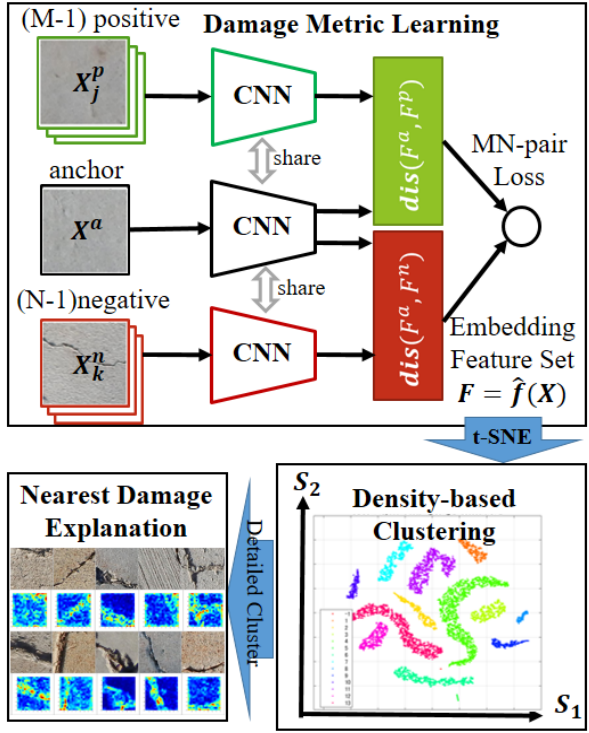

Figure 2. Our pipeline for damage representation learning, embedded clustering, and explanation

Here, this is divided by a normalized temperature scale τ. This scale enhances small values, to ensure that N-pair loss is able to train efficiently, for example, we can set the scale τ=0.3. Thus, the N-pair loss in (2) can be expressed as follows:

$$= -\log \frac{\exp(s_{i,j}^{a,+})}{\exp(s_{i,j}^{a,+}) + \sum_{k=1}^{N-1} \exp(s_{i,k}^{a,-})} \quad (4)$$

The N-pair loss is identical to the InfoNCE loss [6][7]. However, N-pair loss is a slow starter, because of the presence of only one positive image toward N-1 negative images. A positive signal is important to bond the inner embedding space around the same class of each anchor.

Thus, we propose an MN-pair weighting loss instead of (2)-(4), in which we create a set of N-1 negative $\{x_k^-\}_{k=1}^{N-1}$ and M-1 positive $\{x_j^+\}_{j=1}^{M-1}$ for every anchor $x_i$:

$$L_{MN-pair}\left(\boldsymbol{\theta}; x_i, \{x_j^+\}_{j=1}^{M-1}, \{x_k^-\}_{k=1}^{N-1}\right) =$$
$$-\log \frac{v \sum_{j=1}^{M-1} \exp(s_{i,j}^{a,+})}{v \sum_{j=1}^{M-1} \exp(s_{i,j}^{a,+}) + w \sum_{k=1}^{N-1} \exp(s_{i,k}^{a,-})} \quad (5)$$

where $v$ is a positive weight, and $w$ is a negative weight constrained by that $v + w = 1$, for example, $v = 0.15$. To train the parameters $\boldsymbol{\theta}$ under a CNN for damage metric learning, we can minimize the MN-pair loss function $L_{MN-pair}$ using a standard optimizer, such as, the Adam.

## 2.2 Density-Based Damage Clustering

Using the damage-embedded feature $\boldsymbol{F}_i$, with $i \in \{1, ..., n\}$ and the dimension $L$, we can reduce its dimension into two axes of scores using the t-SNE algorithm [8]. Several different concepts exist for clusters of damage representation, including, 1) well-separated clusters, 2) center-based clusters within a specified radius, and 3) density-based clusters. Under a two-dimensional damage-embedding space, we can use either a center-based or a density-based approach. The former is based on the distance from neighboring points to the center, such as the K-means [9], and k-nearest neighbor [10]. To provide an effective approach for representing a heterogeneous subdivided region of damage beyond the predefined classes, we used the density-based clustering algorithm (DBSCAN) [11]. The points of the embedded damage features were classified into 1) *core points* in the interior of a dense region, 2) *border points* on the edge of a dense region, and 3) *noise points* in a sparsely occupied region (a noise or background). The DBSCAN algorithm is formally expressed as

1. Label all points as core, border, or noise points.
2. Eliminate noise points.
3. Put an edge between all core points that are within a user-specified distance parameter ε of each other.
4. Make each group of connected core points into a separate cluster.
5. Assign each border point to one of the clusters of its associated core points.

Because we used a density-based definition of a cluster, it is relatively resistant to noise and can handle clusters with arbitrarily damaged shapes. Therefore, DBSCAN can determine numerous damage clusters that cannot be identified using a center-based algorithm, such as k-means. For damage representation clustering, we can set a distance parameter (for example, ε=3) and a minimum number of neighbors for core points (for example, 10).

## 2.3 Nearest Damage Explanation

CNN models with millions of shared parameters achieve satisfactory performance in contrastive deep metric learning. The reasons for this remain unclear despite the impressive performance. Visualization techniques are mainly divided into a masked sampling and an activation mapping approaches. The former includes the occlusion sensitivity [12] and the local interpretable model-agnostic explanations (LIME) [13]. The merits of this approach is that it does not require in-depth knowledge of the network's architecture, and the disadvantage is that it requires iterative computations per image a significant amount of running time for local partitioning, masked sampling, and output prediction. To represent damage metric contrastively, this output is not an image-based prediction output, but an embedding damage feature $\boldsymbol{F}_i, i \in \{1, ..., n\}$ extracted from a CNN output layer for contrastive damage metric learning. Therefore, we selected an activation map approach such as the class activation map (CAM) [14] or gradient-based extension (Grad-CAM) [15]. The weighting feature map of the CAM is ineffective because it limits the global average pooling (GAP) and fully-connected (FC) at the concluding layer of a CNN.

Inspired by the work of [16], we proposed a gradient-based visualization technique for damage metric learning. The final layer of our proposed MN-pair contrastive learning is a damage-embedding space FC sized 1×1× L. Unlike a classification model, we cannot specify any pair of similarities between the anchor, positive, and negative because of the numerous combinations of their pairs. Practically, we proposed a *reduced similarity score* as a *reduction function* to enhance the average powered similarity for a damage explanation as follows:

$$y^S(\boldsymbol{U}) = \sum_{m=1}^{K} (U^m)^2 \quad (6)$$

where $\boldsymbol{U} = (U^1, ..., U^K)$ is an embedded damage feature with $K$ dimensions before the final output layer with a size of $1 \times 1 \times K$ and serves as the *reduction layer*; for example, FC $1 \times 1 \times 128$. To obtain the similarity contrastive localization map $G_{Grad-CAM}^S \in R^{W \times H}$, we can compute the gradient of the reduced similarity score $y^S(\boldsymbol{U})$ with respect to feature map activations $A^m(x, y)$ of the convolutional or ReLU layer serving as a *feature layer*, that is, $\partial y^S / \partial A^m$. Here, $A^m$ has the size of the array $W \times H \times K$, for example, ReLU $20 \times 20 \times 128$. These gradients flowing back are global averages pooled over the width $x$ and height $y$, respectively to obtain the similarity importance weights.

$$\alpha_m^S = \frac{1}{WH} \sum_x \sum_y \frac{\partial y^S}{\partial A^m(x,y)}, m = 1, \dots, K. \quad (7)$$

Thus, this weight $\alpha_m^S$ represents partial linearization of the CNN downstream from feature map activations $A^m$, and captures the similarity importance. We can use a weighted combination of forward activation maps for damage similarity representation learning.

$$G_{Grad-CAM}^S = ReLU\left(\sum_{m=1}^K \alpha_m^S A^m\right). \quad (8)$$

## 3 Applied Results

We conducted experimental studies using our method and applied it to three damage datasets of damage.

### 3.1 Damage Dataset and Training Setup

We implemented experiments using open-accessed datasets of steel product defects (11,638 samples from steel defect detection [17]) and concrete cracks on decks and pavements (12,633 samples from SDNET2018 [18]). Table 1 illustrates each dataset for the experimental studies containing the target damage, the number of predefined classes, and the partitions of the training and test images, respectively.

Table 1. Datasets for experimental studies

| Target damage | Predefined Classes | Number of training/test |
|---|---|---|
| Steel product defect | 5 | 8,148 / 3,490 |
| Concrete crack | 4 | 8,844 / 3,789 |

Table 2. Summarizes the sizes of input damage images and the number of clusters from the results of DBSCAN clustering under the damage embedding space with two dimensions reduced by the t-SNE. Surprisingly, our contrastive damage representation learning provided two or more multiplied clusters toward the number of predefined classes in Table 1.

Table 2. Training image size and clustering results

| Target damage | Input Image Size [h,w] | Density-Based Output of Clusters |
|---|---|---|
| Steel product defect | [256, 256] | 10 |
| Concrete crack | [256, 256] | 13 |

As listed in Table 3, we set the layer type and output shape of the CNN architecture for our experiments. We set up a simple but practical network with 15 layers containing convolution-ReLU-max pooling, and FC. The architecture had neither batch normalization nor skip connections. We set the mini-batch to 128 and the number of iterations to 2,000–5,000 while training the CNN network to obtain a stable loss curve. We used the Adam optimizer with a learning rate of 0.0001, and set the gradient decay factor to 0.9, and set the squared gradient decay factor to 0.99. We used image augmentation with random erasing [21][22]. Ablation studies were conducted on 128, 144, 160, 180, and 200 pixels, the input size was set to 160×160×3. We also set the output dimensions of the damage-embedding space to 16 after conducting ablation studies on 8, 16, and 32. Furthermore, we set the temperature scale to 0.3, compared with the ablation studies of 0.1, 0.2, 0.3, and 0.5. Finally, we set the hyperparameters of the MN-pair damage contrastive learning M=N to 4 and 5, which corresponds to the number of classes in the target dataset. We also set the positive weight to $v = 0.15$, after testing values of 0.1, 0.15, 0.2, and 0.3. For written space restriction, we eliminated these ablation studies where the evaluation index was the overall accuracy from the damage-feature-based classification trained to test the dataset. To compute the gradient-based activation map Grad-CAM, we set the feature layer to Relu4 and set the reduction layer to FC1, as listed in Table 3.

Table 3. Layer type and shape of CNN architecture

| Layer type | Output Shape (S,S,C) | Params # |
|---|---|---|
| Conv1-Relu1 | 160,160,8 | 224 |
| Maxpool1 | 80,80,8 | -- |
| Conv2-Relu2 | 80,80,32 | 2,336 |
| Maxpool2 | 40,40,32 | -- |
| Conv3-Relu3 | 40,40,64 | 18,496 |
| Maxpool3 | 20,20,64 | -- |
| Conv4-Relu4 | 20,20,128 | 73,856 |
| FC1-Relu5 | 1,1,128 | 6,553,728 |
| FC2 | 1,1,16 | 2,064 |
| total | Learnables | 6.6M |

### 3.2 Steel Defect Representation Results

#### 3.2.1 Steel Defect Embedding Clustering

Figure 3 illustrates the results of steel defect clustering using five colors under the damage-embedding space using the t-SNE. The predefined 1st normal class (red), 3rd inclusion (light green), and patch defect (magenta) were subdivided into two or three regions.

Figure 4 shows the results for the density-based 11 clusters using the HSV-based colors from 1 to 11. Note that an outlier exits in the predefined 4th scratch class. We found that the steel damage representation using damage metric learning did not match the predefined five classes of steel defects, as shown in Figure 3. These results analyze 11 clusters in detail; thus the calibrated clusters were density-based and separated independently, as shown in Figure 4.

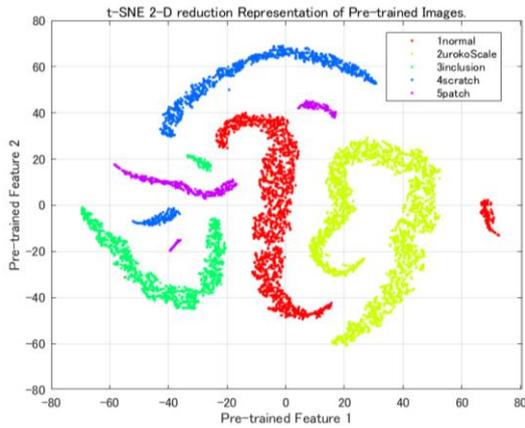

Figure 3. t-SNE visualization of 2-D reduced outputs obtained through training 5 classes of steel defects

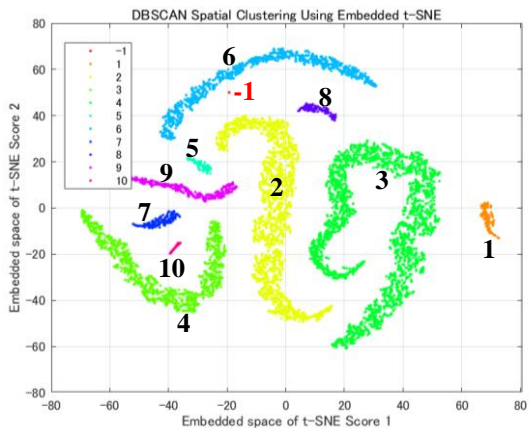

Figure 4. Density-based clustering of Steel product defects into ten using the DBSCAN

#### 3.2.2 Steel Defect of Each Cluster Nearest Images

Figure 5 depicts the density-based ten steel clusters in each row of images, where the first column is sampled, and the nine right images have the nearest similarity score. The numbers on the left-hand side denotes each group shown in Figure 4. The clustered outputs were obtained using a density-based clustering method on a steel dataset.

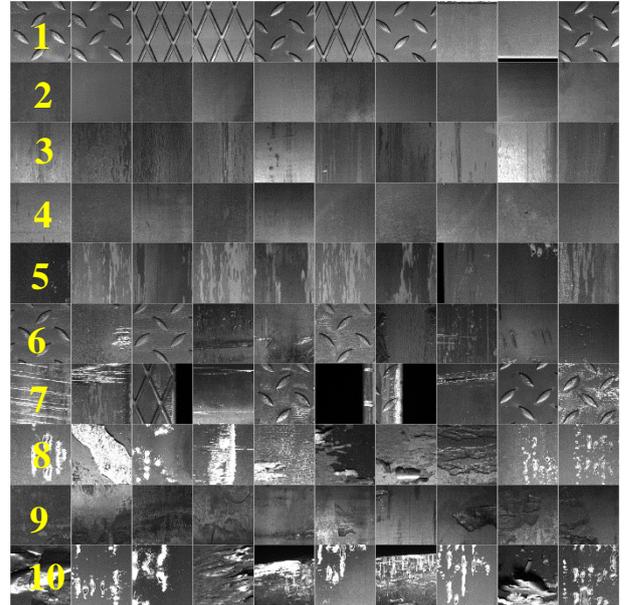

Figure 5. Density-based ten clusters in each row of images, the first column is sampled, and the right nineare nearest images

#### 3.2.3 Steel Product Defect Explanation

Figure 6 illustrates the results of the steel defect explanation by computing the gradient-based activation map using Grad-CAM for contrastive damage metric learning. The order of the rows and columns is the same as that shown in Figure 5. The heat maps illustrate the damage features of the steel defect clusters.

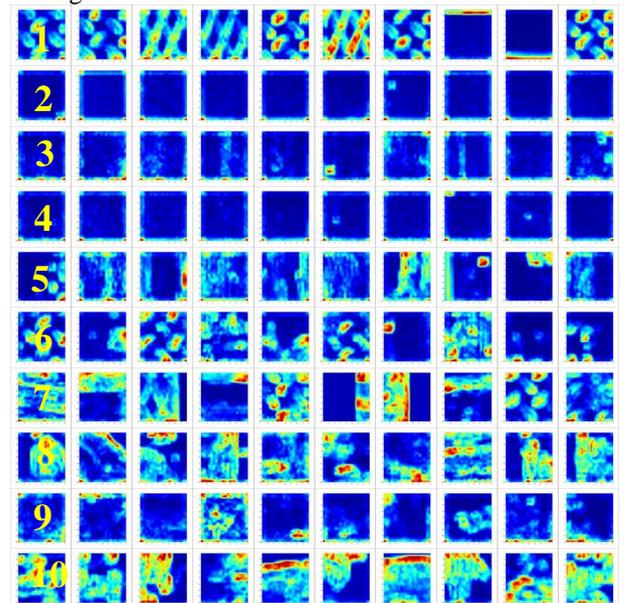

Figure 6. Steel defect of damage explanation

### 3.3 Concrete Crack Representation Results

#### 3.3.1 Concrete Crack Embedding Clustering

Figure 7 illustrates the results of crack clustering using the four colors under the damage-embedding space using t-SNE. The predefined 1st deck-crack class (red), 3rd pavement-crack (light green), and pavement-non-crack (magenta) classes were subdivided into three or four regions.

Figure 8 shows the results for the density-based 14 clusters using HSV-based colors numbered from 1 to 14. Note that one outlier exists in the fourth class of pavements-non-crack. We found that the concrete damage representation using damage metric learning did not match the predefined four classes of cracks, as shown in Figure 7. These results analyze 14 clusters in detail; thus, the calibrated clusters were density-based and separated independently, as shown in Figure 8.

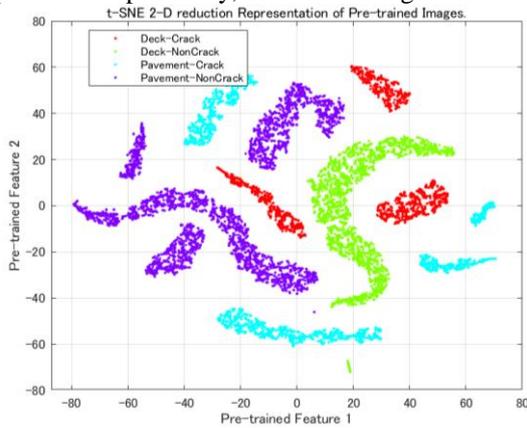

Figure 7. t-SNE visualization of 2-D reduced outputs obtained through training 4 classes of concrete damage

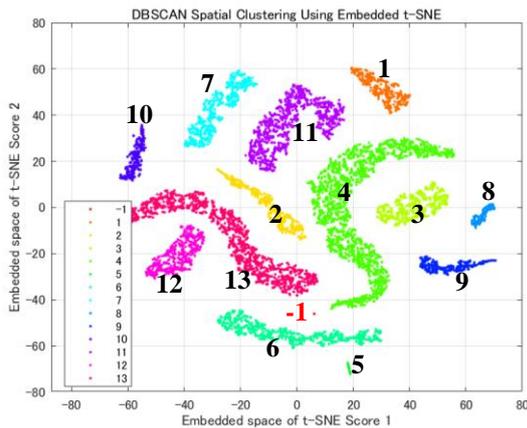

Figure 8. Density-based clustering of Concrete damage into 13 using the DBSCAN

#### 3.3.2 Concrete Crack Clusters Nearest Images

Figure 9 shows the 13 density-based crack clusters in each row image, with the first column sampled and the nine images to the right having the nearest similarity score. Subsequently, the deck and the pavement subgroups were divided.

#### 3.3.3 Concrete Crack Explanation

Figure 10 illustrates the results of the crack explanation by computing the gradient-based activation map using Grad-CAM for contrastive crack metric learning. The arrangement of the rows and columns is consistent with that shown in Figure 9. The heat maps explain the damage features of the crack clusters, and discriminate each background texture between the deck and pavement surfaces. The numbers on the left-hand side denote each group shown in Figure 8. The clustered outputs were obtained using a density-based clustering method on a concrete damage dataset.

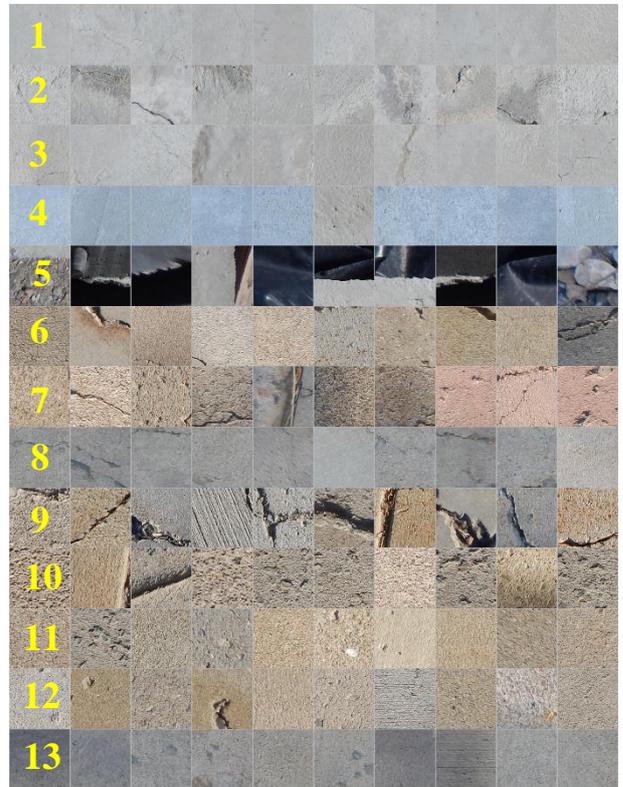

Figure 9. Thirteen density-based clusters in each row of images, the first column is sampled, and the right nine images have nearest similarity scores

## 4 Concluding Remarks

### 4.1 Experimental Results

We proposed a contrastive learning and clustering pipeline for damage representation. Specifically, we introduced the MN-pair damage-embedding space formulation, which includes an anchor and M-1 positive and N-1 negative images. The shared network architecture was a simple but practical convolutional network with 15 layers. Furthermore, we proposed a clustering method using the output of an MN-pair damage-embedding space after two-dimensional-reduction by the t-SNE and density-based DBSCAN clustering. Furthermore, we presented a gradient-based damage explanation to adapt to the final activation map from a convolutional network trained by MN-pair contrastive damage representation learning.

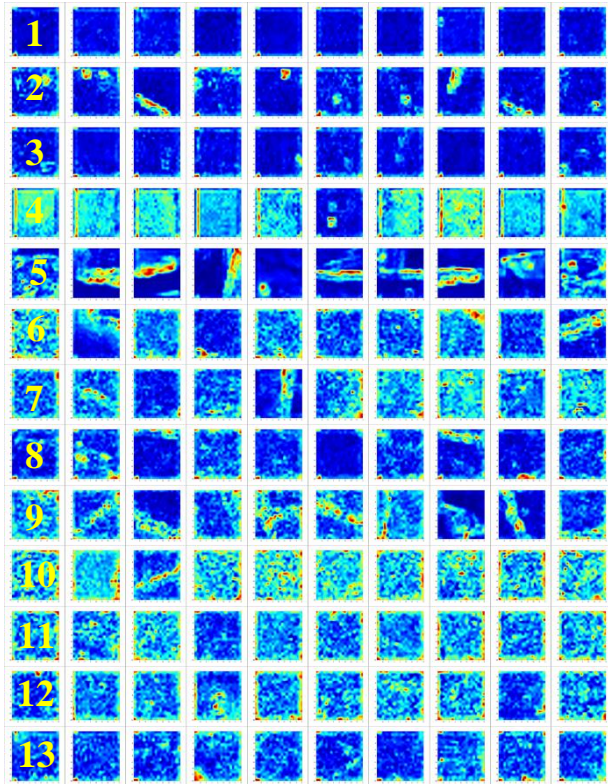

Figure 10. Concrete crack of damage explanation

We found that the damage features exhibited considerable heterogeneous variation compared to the predefined classes determined by human inspection. The damage-embedding space exhibited an unexpected increase in the number of detailed clusters. We demonstrated our method in three experimental studies using the open-access datasets, such as steel products with surface defects, concrete cracks of decks and pavements. We confirmed the efficacy of the proposed method using surface damage datasets.

### 4.2 Limitations and Future Works

We attempted to address the sewer pipe defect dataset, but were hindered by the certain limitations that affected the clarity of the obtained results. Each inspection image was significantly different in terms of the distance and angle toward the damaged region. Simultaneously, the sizes of images were significantly low resolution, with dimensions of 768×576, to clearly visualize the damage region. We believe that a unified distance and angle condition is critical for acquiring inspection images of at least HD quality, that is, 1280×720.

There are more practical problems with *co-defect* representation, where one image has a multiple labels and bordered regions between damaged pairs of classes. To address this issue, a hard-negative sampling scheme was used. For *robustness* against heterogeneous damage, a mixture and erasing augmentation would improve the performance. An MLP-based network is an option for *faster* computation. For *applicability*, other trials should include water quality, disaster scenes, and LCZ built-up/natural classes.

**Acknowledgment** We thank Takuji Fukumoto (MathWorks Japan) for supporting MATLAB resources.